# Cascaded multi-scale and multi-dimension convolutional neural network for stereo matching

Haihua Lu[1], Hai Xu[2], Li Zhang[2], and Yong Zhao[✉]

School of Electronic and Computer Engineering,
Shenzhen Graduate School of Peking University, Shenzhen, China
`luhaihua@pku.edu.cn, yongzhao@pkusz.edu.cn`

**Abstract.** Convolutional neural networks(CNN) have been shown to perform better than the conventional stereo algorithms for stereo estimation. Numerous efforts focus on the pixel-wise matching cost computation, which is the important building block for many start-of-the-art algorithms. However, those architectures are limited to small and single scale receptive fields and use traditional methods for cost aggregation or even ignore cost aggregation. Differently we take them both into consideration. Firstly, we propose a new multi-scale matching cost computation sub-network, in which two different sizes of receptive fields are implemented parallelly. In this way, the network can make the best use of both variants and balance the trade-off between the increase of receptive field and the loss of detail. Furthermore, we show that our multi-dimension aggregation sub-network which containing 2D convolution and 3D convolution operations can provide rich context and semantic information for estimating an accurate initial disparity. Finally, experiments on challenging stereo benchmark KITTI demonstrate that the proposed method can achieve competitive results even without any additional post-processing.

## 1    Introduction

Reconstructing or understanding a 3D scene is crucial in many applications, for instance, autonomous cars, unmanned aerial vehicles and robotics navigation. Although 3D sensors, such as structured light and Lidar, can be employed to capture depth data, utilizing cameras is a more cost-effective solution. Add a constraint that the two input images are a rectified stereo pair of a same scene, then disparity for each pixel in the left image can be computed by matching corresponding pixels on the two images along the horizontal direction [1,2]. However, despite the search space is reduced to 1D, obtaining accurate stereo correspondences is still full of huge challenges for the inherently ill-posed regions, such as textureless, repetitive patterns, occlusions and large saturated areas.

The traditional stereo matching pipeline is divided into four steps: matching cost computation, cost/support aggregation, disparity computation and disparity refinement [3].

Matching cost computation, as a fundamental step in stereo matching algorithms, measures the similarity between two pixels of the two input images. Many conventional stereo algorithms used color intensity, gradient or the distance between pixels as similarity metrics [4,5,6]. Apparently they do not contain enough information, because it is hard to know which features most help measure similarity. In contrast, convolutional neural networks are capable of extracting robust and powerful deep representations directly from the raw data, which have been successful in learning how to match in stereo estimation [7,8,9,10]. Zbontar and LeCun [7] first employed CNN to learn similarity measurement and then used it to initialize the matching cost. Following that, many methods dedicated effort to improving the computation efficiency [11,12] or matching accuracy [13,14]. However these solutions still suffered from several limitations: (i) Limited receptive field. (ii) Single receptive field. (iii) Complex network structure. To break these limitations, we propose a new multi-scale matching cost computation sub-network, in which two different sizes of receptive fields are implemented in parallel. Firstly, since we use pooling and deconvolution operations in the multi-scale sub-network, the model can extract features with a wider receptive field around the target pixels. This allows the model to incorporate more context so that more accurate predictions can be produced in textureless region. On the other hand, the network can make the best use of both variants and balance the trade-off between the increase of receptive field and the loss of detail.

Cost aggregation, dedicating to dealing with mismatching values of the cost volume, is indispensable in traditional local stereo algorithms. Traditionally, it is performed locally by summing/averaging matching cost over windows with constant disparity [15,16,17]. The performances of these methods were limited by the shallow, hand-crafted scheme. To tackle this problem, we propose a learning-based multi-dimension cost aggregation sub-network which containing 2D convolution and 3D convolution operations together. Firstly, we use a 3D convolutional network to effectively incorporate context and geometry information, which is able to operate computation from the height, weight and disparity dimensions [18,19]. After that, we use a 2D convolution network to further improve cost aggregation performance, which provides more semantic information for estimating an accurate initial disparity.

In summary, the contributions of this paper are

- We propose a multi-scale matching cost computation sub-network, in which two different sizes of receptive fields are implemented parallelly. This architecture can balance the trade-off between the increase of receptive field and the loss of detail.
- We present a multi-dimension cost aggregation sub-network which contains 2D convolution and 3D convolution operations together. This model can provide rich context and semantic information for estimating an accurate initial disparity. To the best of our knowledge, this work is the first to joint multi-dimension convolution operations for cost aggregation.
- Experiments show that our architecture can achieve accurate results even without any additional post-processing.

The rest of our paper is organized as follows. Section 2 reviews related work. In section 3, we introduce our model and detail the components. Finally, the experimental results are given in section 4.

## 2 Related work

There are a lot of studies on stereo matching. Here we only review the work that focused on matching cost computation and cost aggregation, which most relevant to our work.

**Matching cost computation.** In traditional stereo algorithms, the absolute gradient differences [20] and absolute intensity differences [21] are the most common pixel-based matching costs. These hand-crafted matching cost metrics lack of sufficient information and robustness. In the contrary, deep learning models can learn more robust and discriminative features that help improve matching cost. Zbontar et al. [7] first employed CNN to learn similarity measurement between two image patches and used it to initialize the matching cost. Luo et al. [11] improved the computation efficiency by learning a probability distribution over all disparity levels under consideration. Chen et al. [12] presented a multi-scale ensemble framework for good local matching scores. Shaked et al. [13] used a highway network with multilevel weighted residual shortcuts for matching cost computation. Patrick et al. [14] proposed a siamese network with pooling and deconvolution operations for similarity computation with a wider receptive field. Our multi-scale matching cost computation sub-network is most similar to that developed by Patrick et al. [14]. The biggest difference is that two different sizes of receptive fields are implemented parallelly in our network. This enables the model to make the best use of both variants and balance the trade-off between the increase of receptive field and the loss of detail. Besides, different from [14] directly upsampling, we concatenate deconvolution features with corresponding feature maps from the encoder part, which enables the model to preserve both the high-level coarser information and low-level fine information.

**Matching cost aggregation.** In general, it is performed locally by summing/averaging matching cost over windows with constant disparity [15,16,17,22]. Yoon et al. [15] used a fixed window filter with adaptive support-weight of each pixel. Zhang et al. [22] used an adaptive window to guarantee the neighbors of each pixel are only from same object. All of these methods based on hand-designed functions and were unable to capture useful context and semantic information, which leaded to their limited performance. Jeong et al. [23] used CNN to learn the convolution kernel for cost aggregation. However, this method need to combine with edge detection task and global energy minimization to achieve a better result. Besides, the aggregation network can't be trained with cost computation network together. In this paper, we propose a multi-dimension aggregation sub-network containing 2D convolution and 3D convolution operations. On the basis of using a 3D convolution network to incorporate rich context and geometry information, a 2D convolution network is used to capture more semantic information to further improve cost aggregation performance.

Our multi-scale matching cost computation sub-network and multi-dimension aggregation sub-network can be trained together. Even without using any post-processing and regularization, the proposed method can estimate an accurate initial disparity.

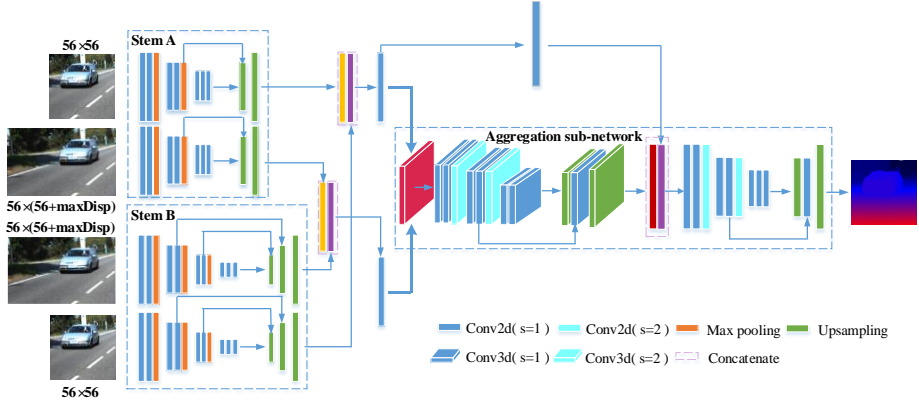

**Fig. 1.** Our network architecture. Stem A and Stem B with different sizes of receptive fields make up the multi-scale matching cost computation sub-network. The aggregation sub-network contains 3D convolution and 2D convolution operations together. The matching cost computation sub-network outputs two cost volumes, which are concatenated as input to the aggregation sub-network. The initial disparity is obtained from the cost aggregation result by a winner-take-all strategy.

## 3 Proposed method

Our network architecture is described in Figure 1, which consists of two sub-networks. The matching cost computation sub-network has two stems with different scales of receptive fields, as shown in Stem A and Stem B. The following sub-network containing 2D convolution and 3D convolution operations together carries out cost aggregation. The more detailed implementations of them are introduced in Table 1 and Table2, respectively.

### 3.1 Multi-scale matching cost computation sub-network

Matching cost computation measures the similarity between two pixels of the two input images. It is designed to construct a cost volume, which represents how well a pixel in the reference image matches the same pixel in the target image shifted by each disparity level under consideration. To compute the matching cost, we first learn deep unary features, and then use them to form a cost volume.

**Unary Features.** Intuitively, the bigger the receptive field is, the more context the model gets. Commonly, pooling, convolution and dilation convolution are the most popular methods to enlarge receptive field. It has been turned out to be that the details lost in pooling operations can be compensated with deconvolution operations as long as these are carried out before computing correlation [14]. Song et al. [24] also argued that the pooling operations perform the best in their context pyramids architecture. Therefore, we also follow this in our work. We use a new multi-scale sub-network to extract the deep representation of each pixel in the reference image and the target image. This sub-network consists of two stems, in which each stem uses pooling and deconvolution operations to get wider receptive field, as shown in Figure 1. Since the structure of the two stems is the same, stem A is taken as an example for detailed description here.

Stem A is an siamese network, which consists of two shared-weight sub-networks to handle the two input images simultaneously. Each sub-network of the siamese network is an encoder-decoder network, as described in Table 1. The encoder part consists of seven 3×3 convolution layers in series, where each followed by a batch normalized layer and a rectified linear unit except the last one. In addition to the last three layers, there is a max pooling layer between every two convolution layers. Then, the decoder part implements the same number of 3×3 deconvolution layers as the max pooling layers. Different from [14] directly upsampling, we concatenate deconvolution features with corresponding feature maps from the encoder part, which enables the model to preserve both the high-level coarse information and low-level fine information. The structure of stem B is the same as that of stem A. The only difference is that stem B has a deeper structure with more max pooling layers. In this case, stem B has a wider receptive field, but loses more detail.

In order to balance the trade-off between the increase of receptive field and the loss of detail, our model implements the two stems in parallel to obtain different scales of receptive fields. In this way, the network could make the best use of them.

**Cost volume.** After getting deep unary features, the next step is to compute the stereo matching cost. We use the unary features from the two stems to construct a cost volume. Firstly, we concatenate the features from the left branch of stem A with the features from the left branch of stem B and use a 1×1 convolution layer to obtain the final left unary features. Similarly, the features from the right branch of stem A and the features from right branch of stem B are also concatenated to get the final right unary features by using a 1×1 convolution layer. After that, as the way [18] proposed, each unary feature from the left unary features is concatenated with the corresponding unary feature from the right unary features across every disparity level under consideration. Finally, we get a 4D cost volume with size of *(max disparity+1) × height × width × feature channel* by packing the concatenating results.

### 3.2 Multi-dimension Cost aggregation sub-network

Even if the deep unary features are used, mismatching still exists in the cost volume. Hence, cost aggregation is indispensable in refining these representations. We firstly use a 3D convolution network to take into account the rich context and geometry information in the cost volume, and then incorporate deep semantic information by a 2D convolution network to further improve cost aggregation performance.

**3D convolution network for context learning.** 3D convolutional network is able to operate computation from the height, weight and disparity dimensions, which has great potential to capture context and geometry information [18,19]. But the biggest problem is that 3D convolutions bring computation and memory burden. An effective solution is to use an encoder-decoder structure, which also enables the model to take advantage of more context with a wider receptive field.

**Table 1.** The detailed implementation of multi-scale matching cost computation sub-network

| Layer | K | S | Ch(I/O) | OutRes | Input |
|---|---|---|---|---|---|
| **Stem A** | | | | | |
| conv_a1 | 3×3 | 1 | 3/32 | H×W | imageL, imageR |
| conv_a2 | 3×3 | 1 | 32/32 | H×W | conv_a1 |
| pool_a1 | 2×2 | 2 | 32/32 | 1/2H×1/2W | conv_a2 |
| conv_a3 | 3×3 | 1 | 32/32 | 1/2H×1/2W | pool_a1 |
| conv_a4 | 3×3 | 1 | 32/32 | 1/2H×1/2W | conv_a3 |
| pool_a2 | 2×2 | 2 | 32/32 | 1/4H×1/4W | conv_a4 |
| conv_a5 | 3×3 | 1 | 32/32 | 1/4H×1/4W | pool_a2 |
| conv_a6 | 3×3 | 1 | 32/32 | 1/4H×1/4W | conv_a5 |
| conv_a7 | 3×3 | 1 | 32/32 | 1/4H×1/4W | conv_a6 |
| upconv_a1 | 3×3 | 2 | 32/32 | 1/2H×1/2W | conv_a7 |
| upconv_a2 | 3×3 | 2 | 32/32 | H×W | upconv_a1, conv_a4 |
| **Stem B** | | | | | |
| conv_b1 | 3×3 | 1 | 3/32 | H×W | imageL, imageR |
| conv_b2 | 3×3 | 1 | 32/32 | H×W | conv_b1 |
| pool_b1 | 2×2 | 2 | 32/32 | 1/2H×1/2W | conv_b2 |
| conv_b3 | 3×3 | 1 | 32/32 | 1/2H×1/2W | pool_b1 |
| conv_b4 | 3×3 | 1 | 32/32 | 1/2H×1/2W | conv_b3 |
| pool_b2 | 2×2 | 2 | 32/32 | 1/4H×1/4W | conv_b4 |
| conv_b5 | 3×3 | 1 | 32/32 | 1/4H×1/4W | pool_b2 |
| conv_b6 | 3×3 | 1 | 32/32 | 1/4H×1/4W | conv_b5 |
| pool_b3 | 2×2 | 2 | 32/32 | 1/8H×1/8W | conv_b6 |
| conv_b7 | 3×3 | 1 | 32/32 | 1/8H×1/8W | pool_b3 |
| conv_b8 | 3×3 | 1 | 32/32 | 1/8H×1/8W | conv_b7 |
| conv_b9 | 3×3 | 1 | 32/32 | 1/8H×1/8W | conv_b8 |
| upconv_b1 | 3×3 | 2 | 32/32 | 1/4H×1/4W | conv_b9 |
| upconv_b2 | 3×3 | 2 | 32/32 | 1/2H×1/2W | upconv_b1, conv_b6 |
| upconv_b3 | 3×3 | 2 | 32/32 | H×W | upconv_b2, conv_b4 |
| **Cost volume** | | | | | |
| conv_1 | 1×1 | 1 | 64/32 | H×W | featureL_a, featureL_b |
| conv_2 | 1×1 | 1 | 64/32 | H×W | featureR_a, featureR_b |
| cost-volume | - | - | -/64 | D×H×W | conv_1, conv_2 |

As described in Table 2, the cost volume is first handled by eight 3×3×3 3D convolution layers in series, where each followed by a batch normalized layer and a rectified linear unit. There is a 3D convolution layers with stride 2 between every two 3D convolution layers with stride 1. We sub-sample the cost volume twice by a factor of 4 in the encoder part. Similarly, for the decoder part, we up-sample the cost volume twice and add the corresponding same resolution feature maps from the encoder part before up-sampling. The additional convolution layer in the decoder part can produce smoother disparity [9]. At the end of the 3D convolution network, we get a regularized cost volume with size of *(max disparity+1) × height × width×1*.

**2D convolution network for semantic information learning.** On the basis of using a 3D convolution network to incorporate context and geometry information, we use another 2D convolution network to capture more semantic information to further improve cost aggregation performance. Firstly, we transpose the cost volume from the 3D convolution network into a new cost volume with size of *height × width × (max disparity+1)*. Then we concatenate it with the low-level features from the matching cost computation sub-network in order to incorporate low-level structure information. After that, we use a 2D convolution network which has the same structure as 3D convolution network to get a final cost volume.

**Table 2.** The detailed implementation of multi-dimension cost aggregation sub-network. Unless otherwise specified, the convolution stride for each dimension is the same.

| Layer | K | S | Ch(I/O) | OutRes | Input |
|---|---|---|---|---|---|
| **Agg_3D** | | | | | |
| conv3d_1 | 3×3×3 | 1 | 64/16 | D ×H×W | cost-volume |
| conv3d_2 | 3×3×3 | 1 | 16/16 | D ×H×W | conv3d_1 |
| conv3d_3 | 3×3×3 | 2 | 16/32 | 1/2D×1/2H×1/2W | conv3d_2 |
| conv3d_4 | 3×3×3 | 1 | 32/32 | 1/2D×1/2H×1/2W | conv3d_3 |
| conv3d_5 | 3×3×3 | 1 | 32/32 | 1/2D×1/2H×1/2W | conv3d_4 |
| conv3d_6 | 3×3×3 | 2 | 32/64 | 1/4D×1/4H×1/4W | conv3d_5 |
| conv3d_7 | 3×3×3 | 1 | 64/64 | 1/4D×1/4H×1/4W | conv3d_6 |
| conv3d_8 | 3×3×3 | 1 | 64/64 | 1/4D×1/4H×1/4W | conv3d_7 |
| upconv3d_1 | 3×3×3 | 2 | 64/32 | 1/2D×1/2H×1/2W | conv3d_8 |
| conv3d_9 | 3×3×3 | 1 | 32/32 | 1/2D×1/2H×1/2W | upconv3d_1+conv3d_5 |
| upconv3d_2 | 3×3×3 | 2 | 32/1 | D ×H×W | conv3d_9 |
| **Agg_2D** | | | | | |
| conv2d_1 | 1×1 | 1 | 32/16 | H×W | featureL(conv_1) |
| transpose | - | - | -/129 | H×W | upconv3d_2 |
| conv2d_2 | 3×3 | 1 | 145/16 | H×W | conv2d_1, transpose |
| conv2d_3 | 3×3 | 1 | 16/16 | H×W | conv2d_2 |
| conv2d_4 | 3×3 | 2 | 16/32 | 1/2H×1/2W | conv2d_3 |
| conv2d_5 | 3×3 | 1 | 32/32 | 1/2H×1/2W | conv2d_4 |
| conv2d_6 | 3×3 | 1 | 32/32 | 1/2H×1/2W | conv2d_5 |
| conv2d_7 | 3×3 | 2 | 32/64 | 1/4H×1/4W | conv2d_6 |
| conv2d_8 | 3×3 | 1 | 64/64 | 1/4H×1/4W | conv2d_7 |
| conv2d_9 | 3×3 | 1 | 64/64 | 1/4H×1/4W | conv2d_8 |
| upconv2d_1 | 3×3 | 2 | 64/32 | 1/2H×1/2W | conv2d_9 |
| conv2d_10 | 3×3 | 1 | 64/32 | 1/2H×1/2W | upconv2d_1, conv2d_6 |
| upconv2d_2 | 3×3 | 2 | 32/129 | H×W | conv2d_10 |
| WTA | - | - | -/1 | H×W | upconv2d_2 |

Finally, the initial disparity $D(\boldsymbol{p})$ is obtained from the cost aggregation result $C(\boldsymbol{p}, d)$ by a winner-take-all(WTA) strategy:

$$D(\boldsymbol{p}) = \operatorname*{argmin}_{d} C(\boldsymbol{p}, d)$$

Where $\boldsymbol{p}$ is a pixel of the reference image, $d$ is the disparity level under consideration.

## 4    Experimental results

In this section, we evaluate our model on KITTI 2012 [27] and KITTI 2015 [28] dataset. Our architecture is implemented by the Tensorflow [29] with a AdaGrad optimization method [30] and a constant learning rate 0.001. Prior to training, each input image is normalized to zero mean and standard deviation of one. We adopt a batch size of 8, using a H×W randomly located crop from the reference image and a H × (W+ D) crop from the target image. We use H=58, W=58, D=128(max disparity) in our experiments. We trained the network for about 400k iterations which takes around 2 days on a single NVIDIA 1080Ti GPU.

**Table 3.** Comparisons of the output of the matching network on the KITTI 2012 validation set.

| model | >2 px | | >3 px | | >5 px | | End-Point | |
|---|---|---|---|---|---|---|---|---|
| | Non-Occ | All | Non-Occ | All | Non-Occ | All | Non-Occ | All |
| MC-CNN-acrt | 15.02 | 16.92 | 12.99 | 14.93 | 11.38 | 13.32 | 4.39px | 5.21px |
| MC-CNN-fast | 17.72 | 19.56 | 15.53 | 17.41 | 13.60 | 15.51 | 4.77px | 5.63px |
| Content-CNN(19) | 10.87 | 12.86 | 8.61 | 10.64 | 7.00 | 9.03 | 3.31px | 4.20px |
| Our(P2) | 9.11 | 11.12 | 7.33 | 9.38 | 5.71 | 7.76 | 2.16px | 3.05px |
| Our(P3) | 9.30 | 11.35 | 7.42 | 9.51 | 5.68 | 7.79 | 2.23px | 3.16px |
| Our(final) | **8.31** | **10.32** | **6.50** | **8.54** | **4.81** | **6.88** | **1.83px** | **2.65px** |

**Table 4.** Comparisons of the output of the matching network on the KITTI 2015 validation set.

| model | >2 px | | >3 px | | >5 px | | End-Point | |
|---|---|---|---|---|---|---|---|---|
| | Non-Occ | All | Non-Occ | All | Non-Occ | All | Non-Occ | All |
| MC-CNN-acrt | 15.20 | 16.83 | 12.45 | 14.12 | 10.13 | 11.80 | 4.01px | 4.66px |
| MC-CNN-fast | 18.47 | 20.04 | 14.96 | 16.59 | 12.02 | 13.67 | 4.27px | 4.93px |
| Content-CNN(37) | 9.96 | 11.67 | 7.23 | 8.97 | 5.04 | 6.78 | 1.84px | 2.56px |
| Our(P2) | 9.89 | 10.58 | 6.32 | 8.73 | 4.45 | 6.09 | 1.85px | 2.50px |
| Our(P3) | 8.66 | 10.23 | 6.01 | 7.64 | 4.21 | 5.82 | 1.81px | 2.46px |
| Our(final) | **7.65** | **9.23** | **5.54** | **7.12** | **3.93** | **5.48** | **1.50px** | **2.00px** |

### 4.1 Matching cost computation evaluation

To demonstrate the effectiveness of the multi-scale matching cost computation sub-network, we compare it to existing matching networks [7,11]. In this experiment, we use an inner-product operation to compute the cost volume. Table 3 shows the comparisons on KITTI 2012 validation set and Table 4 describes the comparisons on KITTI 2105 validation set. The results show that our single stem networks(P2, P3) can outperform previous matching networks with a large margin, which demonstrates the advantages of pooling/deconvolution encoder-decoder network. Besides, we can see that the multi-scale network(final) works better than the single stem networks. This testifies that parallelly implementing two sizes of receptive fields can make the best use of them and balance the trade-off between the increase of receptive field and the loss of detail.

### 4.2 Matching cost aggregation evaluation

**Table 5.** Non-occluded 3-pixel error on KITTI 2012 validation set with different cost aggregation method.

| Unary | Cost Aggregation | | | Our(*) | Content-CNN(+) |
|-------|-------|-------|------|--------|----------------|
| | conv3d | conv2d | CBCA | | |
| *+ | | | | 6.50 | 6.61 |
| *+ | * | | | 4.93 | - |
| *+ | | * | | 4.72 | - |
| *+ | * | * | + | **3.69** | 6.09 |

**Table 6.** Non-occluded 3-pixel error on KITTI 2015 validation set with different cost aggregation method.

| Unary | Cost Aggregation | | | Our(*) | Content-CNN(+) |
|-------|-------|-------|------|--------|----------------|
| | conv3d | conv2d | CBCA | | |
| *+ | | | | 5.54 | 7.13 |
| *+ | * | | | 3.95 | - |
| *+ | | * | | 3.80 | - |
| *+ | * | * | + | **3.01** | 6.58 |

Experimental results of [23] have proved that the conventional hand-crafted aggregation method CBCA [4] outperforms the other traditional aggregation methods such as BF [3], GF [31] and DT [32]. So we compare the performance of our multi-dimension matching cost aggregation network to CBCA in this section. For fair comparisons, we compare to the results of [11] with only CBCA. From Table 6, we can see that CBCA lowered the KITTI 2015 validation error of Content-CNN(+) from 7.13 to 6.58, which just had a slight improvement with 0.55. In contrast, our method(*) reduces the error from 5.54 to 3.01 by a large margin of 2.53. It demonstrates that our multi-dimension cost aggregation sub-network performs significantly better than the conventional hand-crafted aggregation methods. Table 5 shows the same results on KITTI 2012 validation set.

On the other hand, we also compare our different model variants. Both Table 5 and Table 6 prove that the error can be effectively reduced using only the 3D convolution network or the 2D convolution network. When cascading the two networks together, we can get even greater improvements.

### 4.3 Benchmark results

We evaluate our final results on KITTI 2012 [27] and KITTI 2015 [28] benchmarks, which consist of challenging and complicated road scenes captured by driving. The KITTI 2012 dataset contains 194 training and 195 testing images, and the KITTI 2015 dataset consists of 200 training and 200 testing images. The ground truths were obtained from LIDAR data.

Table 7 shows the comparisons on the KITTI 2012 benchmark. We mainly compare our model to those methods which leveraged deep learning representations to compute matching cost only and used several traditional regularization and post-processing steps to refine their results, such as MC-CNN [7], Content-CNN [11], Deep Embed [12]. Strikingly, our model achieves on-par results even without using any post-processing and regularization. For non-occlusion, our results even exceed them. Besides, we also compare with other cost aggregation methods including CAT [33] and S+GF [4].

**Table 7.** Comparisons on KITTI 2012 benchmark

| Model | >2 px | | >3 px | | Mean Error | | Runtime(s) |
|---|---|---|---|---|---|---|---|
| | Non-Occ | All | Non-Occ | All | Non-Occ | All | |
| GC-NET [18] | **2.71** | **3.46** | **1.77** | **2.30** | **0.6px** | **0.7px** | 0.9 |
| MC-CNN-acrt [7] | 3.90 | 5.45 | 2.43 | 3.63 | 0.7px | 0.9px | 67 |
| Content-CNN [11] | 4.98 | 6.51 | 3.07 | 4.29 | 0.8px | 1.0px | 0.7 |
| DispNetC [9] | 7.38 | 8.11 | 4.11 | 4.65 | 0.9px | 1.0px | **0.06** |
| Deep Embed [12] | 5.05 | 4.67 | 3.10 | 4.24 | 0.9px | 1.1px | 3 |
| SPS-st [26] | 4.98 | 6.28 | 3.39 | 4.41 | 0.9px | 1.0px | 2 |
| CAT [33] | 8.11 | 9.44 | 3.31 | 4.07 | 1.1px | 1.2px | 10 |
| S+GF [4] | 14.72 | 16.76 | 5.53 | 7.79 | 2.1px | 3.4px | 140 |
| Ours | 3.91 | 5.89 | 2.62 | 4.50 | 1.0px | 1.6px | 1.2 |

**Table 8.** Comparisons on KITTI 2015 benchmark

| Model | All pixels | | | Non-Occluded pixels | | | Runtime(s) |
|---|---|---|---|---|---|---|---|
| | D1-bg | D1-fg | D1-all | D1-bg | D1-fg | D1-all | |
| GC-NET [18] | **2.21** | 6.16 | **2.87** | **2.02** | 5.58 | **2.61** | 0.9 |
| MC-CNN-acrt [7] | 2.89 | 8.88 | 3.89 | 2.48 | 7.64 | 3.33 | 67 |
| Content-CNN [11] | 3.73 | 8.58 | 4.54 | 3.32 | 7.44 | 4.00 | 1 |
| DispNetC [9] | 4.32 | **4.41** | 4.34 | 4.11 | **3.72** | 4.05 | **0.06** |
| 3DMST [25] | 3.36 | 13.03 | 4.97 | 3.03 | 12.11 | 4.53 | 93 |
| SPS-st [26] | 3.84 | 12.67 | 5.31 | 3.50 | 11.61 | 4.84 | 2 |
| Ours | 4.06 | 8.67 | 4.59 | 3.01 | 6.59 | 3.40 | 1.2 |

Table 8 shows the comparisons on the KITTI 2015 benchmark, which demonstrates the same results as KITT 2012 benchmark.

Qualitative results are depicted in Figure 2. We can observe that our method performs well in the textureless regions, for instance, the road. However, it suffers from the occlusion regions where pixels appearing in the reference image are occluded in the target image. This is due to we do not use any post-processing and regularization. We leave this problem of designing a robust post-processing module for future investigations.

## 5    Conclusions

We successfully demonstrate a method that can achieve competitive results on KITTI 2012 and KITTI 2015 benchmarks even without any additional post-processing and regularization. Firstly, a new multi-scale sub-network for matching cost computation is proposed, in which two different sizes of receptive fields are implemented parallelly to balance the trade-off between the increase of receptive field and the loss of detail. Then, a multi-dimension aggregation sub-network containing 2D convolution and 3D convolution operations is designed to provide rich context and semantic information for estimating an accurate initial disparity.

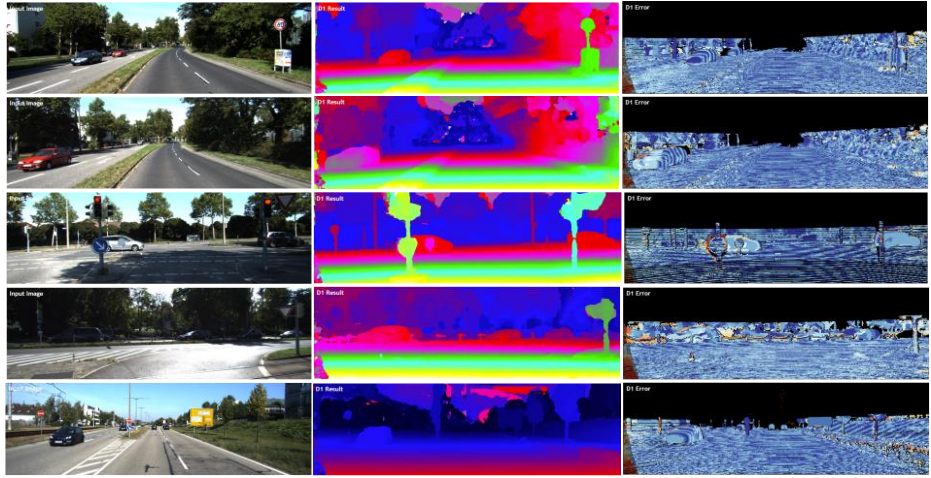

**Fig. 2.** Qualitative results on KITTI2015 benchmark. From left: left stereo input image, disparity prediction, error map.

In our present work, we only focus on the matching cost computation and cost aggregation steps in the pipeline for stereo matching without using any additional post-processing and regularization. Therefore, our next goal is to design a robust post-processing module which will contribute to an accurate disparity.

## References

1. Barnard, S., Fischler, M.: Computational stereo. ACM Computing Surveys 14, 553-572 (1982).
2. Brown, M. Z., Burschka, D., Hager, G. D.: Advances in computational stereo. IEEE Transactions on Pattern Analysis & Machine Intelligence 25(8), 993-1008 (2003).
3. Scharstein, D., Szeliski, R., Zabih, R.: A Taxonomy and Evaluation of Dense Two-Frame Stereo Correspondence Algorithms. International Journal of Computer Vision 47(1-3), 7-42 (2002).
4. Zhang, K., Fang, Y., Min, D., Sun, L., Yang, S., Yan, S.: Cross-Scale Cost Aggregation for Stereo Matching. In IEEE Conference on Computer Vision and Pattern Recognition. pp. 1590-1597 (2014).
5. Mei, X., Sun, X., Dong, W., Wang, H., Zhang, X.: Segment-Tree Based Cost Aggregation for Stereo Matching. IEEE International Conference on Acoustics, Speech and Signal Processing. vol. 9, pp. 313-320 (2017).
6. Yang, Q.: A non-local cost aggregation method for stereo matching. In IEEE Conference on Computer Vision and Pattern Recognition. pp. 1402-1409 (2012).
7. Zbontar, J., LeCun, Y.: Stereo matching by training a convolutional neural network to compare image patches. Journal of Machine Learning Research 17(1), 2287-2318 (2016).

8.  Liang, Z., Feng, Y., Guo, Y., Qiao, L.: Learning Deep Correspondence through Prior and Posterior Feature Constancy. (2017).
9.  Mayer, N., Ilg, E., Hausser, P., Fischer, P., Cremers, D., Dosovitskiy, A., Brox, T.: A large dataset to train convolutional networks for disparity, optical flow, and scene flow estimation. In IEEE Conference on Computer Vision and Pattern Recognition. pp. 4040-4048 (2016).
10. Gidaris, S., Komodakis, N.: Detect, replace, refine: Deep structured prediction for pixel wise labeling. In IEEE Conference on Computer Vision and Pattern Recognition. pp. 7187-7196 (2017).
11. Luo, W., Schwing, A. G., Urtasun, R.: Efficient deep learning for stereo matching. In IEEE Conference on Computer Vision and Pattern Recognition. pp. 5695-5703 (2016).
12. Chen, Z., Sun, X., Wang, L., Yu, Y., Huang, C.: A Deep Visual Correspondence Embedding Model for Stereo Matching Costs. IEEE International Conference on Computer Vision. pp. 972-980 (2016).
13. Shaked, A., Wolf, L.: Improved stereo matching with constant highway networks and reflective confidence learning. IEEE Conference on Computer Vision and Pattern Recognition. pp. 6901-6910 (2017).
14. Brandao, P., Mazomenos, E., Stoyanov, D.: Widening siamese architectures for stereo matching (2017).
15. Yoon, K., Kweon, I.: Adaptive support-weight approach for correspondence search. IEEE Trans Pattern Anal Mach Intell 28(4), 650–656 (2006).
16. Pham, C. C., Jeon, J. W.: Domain Transformation-Based Efficient Cost Aggregation for Local Stereo Matching. IEEE Transactions on Circuits & Systems for Video Technology 23(7), 1119-1130 (2013).
17. Hosni, A., Rhemann, C., Bleyer, M., Rother, C., Gelantz, M.: Fast Cost-Volume Filtering for Visual Correspondence and Beyond. IEEE transactions on pattern analysis and machine intelligence 35(2), 504-11 (2013).
18. Kendall, A., Martirosyan, H., Dasgupta, S., Henry, P., Kennedy, R., Bachrach, A., Bry, A.: End-to-end learning of geometry and context for deep stereo regression (2017).
19. Yu, L., Wang, Y., Wu, Y., Jia, Y.: Deep Stereo Matching with Explicit Cost Aggregation Sub-Architecture. (2018).
20. Scharstein, D.: Matching images by comparing their gradient fields. Iapr International Conference on Pattern Recognition - Conference A: Computer Vision & Image Processing. vol. 1, pp. 572-575 (1994).
21. Kanade, T., Kano, H., Kimura, S., Yoshida, A. Oda, K.: Development of a video-rate stereo machine. IEEE/RSJ International Conference on Intelligent Robots and Systems 95. 'human Robot Interaction and Cooperative Robots'. vol. 3, pp. 95-100 (1995).
22. Zhang, K., Lu, J., Lafruit, G.: Cross-based local stereo matching using orthogonal integral images. IEEE Transactions on Circuits & Systems for Video Technology 19(7), 1073-1079 (2009).
23. Jeong, S., Kim, S., Ham, B., Sohn, K.: Convolutional cost aggregation for robust stereo matching. IEEE International Conference on Image Processing. pp. 2523-2527 (2017).
24. Song, X., Zhao, X., Hu, H. W., Fang, L.: EdgeStereo: a context integrated residua pyramid network for stereo matching. (2018).
25. Li, L., Yu, X., Zhang, S., Zhao, L., Zhang, L.: 3D cost aggregation with multiple minimum spanning trees for stereo matching. Applied Optics 56(12), 3411-3420 (2017).
26.  Yamaguchi, K., Mcallester, D., Urtasun, R.: Efficient Joint Segmentation, Occlusion Labeling, Stereo and Flow Estimation. European Conference on Computer Vision. pp. 756-771 (2014).
27. Geiger, A.: Are we ready for autonomous driving? The KITTI vision benchmark suite. Computer Vision and Pattern Recognition. pp. 3354-3361 (2012).

28. Menze, M., Geiger, A.: Object scene flow for autonomous vehicles. Computer Vision and Pattern Recognition. pp. 3061-3070 (2015).
29. Abadi, M., Agarwal, A., Barham, P., Brevdo, E., Chen, Z., Citro, C., Corrado, G. S., Davis, A., Dean, J., Devin, M., et al.: Tensorflow: Largescale machine learning on heterogeneous distributed systems. (2016).
30. Duchi, J., Hazan, E., Singer, Y.: Adaptive Subgradient Methods for Online Learning and Stochastic Optimization. Journal of Machine Learning Research 12(7), 257-269 (2011).
31. Tang, X.: Guided image filtering. IEEE Transactions on Pattern Analysis & Machine Intelligence 35(6), 1397-409 (2010).
32. Gastal, E. S. L., Oliveira, M. M.: Domain transform for edge-aware image and video processing. ACM Transactions on Graphics 30(4), 1-12 (2011).
33. Ha, J. M., Jeon, J. Y., Bae, G. Y., Jo, S. Y., Hong, J.: Cost Aggregation Table: Cost Aggregation Method Using Summed Area Table Scheme for Dense Stereo Correspondence. Advances in Visual Computing. Springer International Publishing. pp. 815-826 (2014).